# The Topological Fusion of Bayes Nets *

**Izhar Matzkevich**  **Bruce Abramson**
Computer Science Department and Social Science Research Institute
University of Southern California
Los Angeles, CA, 90089-0781
izhar@pollux.usc.edu
bda@pollux.usc.edu

## Abstract

Bayes nets are relatively recent innovations. As a result, most of their theoretical development has focused on the simplest class of *single-author* models. The introduction of more sophisticated *multiple-author* settings raises a variety of interesting questions. One such question involves the nature of compromise and consensus. Posterior compromises let each model process all data to arrive at an independent response, and then split the difference. Prior compromises, on the other hand, force compromise to be reached on all points *before* data is observed. This paper introduces prior compromises in a Bayes net setting. It outlines the problem and develops an efficient algorithm for fusing two directed acyclic graphs into a single, consensus structure, which may then be used as the basis of a prior compromise.

## 1 INTRODUCTION

Bayes nets, belief networks, and influence diagrams are rapidly coming to dominate the work of AI researchers interested in uncertain information, probabilistic relationships, and hierarchical Bayesian inference. These three models, which are variations on a common theme, wed graph theory to decision theory in a fairly straightforward manner: directed acyclic graphs (DAG's) capture qualitative relationships among variables, and mathematical functions (mostly probability distributions) specify the quantitative character of the relationships. The variety of names associated with these models reflects a variety of purposes. Decision analysts interested in the construction of decision models use influence diagrams to capture all of the elements of decision theory— probability, utility, and decision-making—and the relationships among them. Designers of intelligent systems use belief networks to construct knowledge bases containing a domain's objects and their (probabilistic and deterministic) interrelationships. Theoreticians interested in understanding the interplay between probability theory and graph theory study Bayes nets, whose nodes contain propositions of uncertain truth value and whose arcs indicate conditional dependence; dependence, in turn, is specified as conditional distributions.

*Supported in part by the National Science Foundation under grant SES-9106440.

Research on Bayes nets (sometimes extended to influence diagrams) has led to an elegant theoretical foundation for all of these models [8]. Since the models are new, however, the known theory deals only with rather fundamental issues. Although it may be a bit of an overgeneralization, it is probably fair to describe most of the theory of Bayes nets as directed towards understanding the manipulation of information *inside a single network*. The literature on implemented models reflects a similar concern; networks are generally assumed to capture information provided by an individual (or by a group that has agreed upon a common set of opinions) known as the model's *author* [3]. Relatively little has been said about the coordination of information across multiple networks, either in theory or in practice.

Relatively little, however, does not mean *nothing*. Heckerman considered the coordination of multiple belief networks as part of his development of the *similarity network* formalism [2]. In these models—which were developed in a theoretical setting and then implemented in a specific medical system—a large diagnostic problem is partitioned into a set of smaller problems. A distinct local belief network is then devised for each subproblem. Finally, the set of local belief networks is combined into a single global belief network. The fundamental principle guiding this combination is graph union; the global network contains all of the nodes and all of the arcs of all of the local networks. Graph intersection played an important role in Bonduelle's study of the sources of disagreement among experts in a decision making context [1]. His proposed resolution of *model-structure disagreement*, (i.e., structural differences among influence diagrams with different authors), begins by identifying the "core" set of nodes and arcs common to *all* of the input diagrams (or, in other words, their *intersection*). A combination of behavioral and mechanical techniques are used to refine the core to a more meaningful consensus structure. Shachter addressed a somewhat subtler coordination problem, but one that bears a definite kinship to topological fusion [11]. He examined the possibility of imposing a partial ordering on an influence diagram (other than the natural one induced by its arcs), and introduced a (somewhat informal) "bubble sort"-based *interchange algorithm* that used three interchange operations to convert a given initial node sequence into a desired target node sequence. Since this algorithm was developed in a query-answering setting, however, its relationship to the consensus problem was never explored.

This previous work notwithstanding, the problem of coordinating multiple-author networks has received far less than its fair share of attention. The coordination of information across local networks is of obvious importance if the resulting global structure is to remain a valid belief network. Heckerman's choice of a medical domain, and his assumption that all local networks are attributable to the same author, however, restricted the number of coordination issues that he had to consider while developing the similarity network.

Multiple belief networks (by multiple authors) may arise through a variety of circumstances. They may be part of a distributed system-design effort, they may be designed by independent teams initially unaware of each other's existence, they may be local networks that need to be combined into a global network, etc. Regardless of the origin of these multiple networks, some coordination mechanism is necessary. The easiest strategy, of course, is to discard all networks but one. Although this approach is the strategy of choice at times, it is probably most appropriate as a strategy of last resort. A second strategy is to run the individual networks in parallel, and to combine them only after they have each reached a conclusion. This approach, which forms the basis of the probabilistic multi-knowledge-base (PMKBS) architecture, has already shown some encouraging results [6, 7]. A third strategy is to extend the graph union concept introduced with similarity networks by fusing the multiple networks together into a single network with a consensus structure. All three of these strategies are likely to produce different answers.

In this paper, we consider the third option, the fusion of multiple networks into a single one. We make no assumptions about the authorship and/or overlap among these networks, other than the competence and good faith of the authors and the existence of some overlap (coordination of non-overlapping networks is trivial). We also defer discussions of compromise probabilities to a later article; our concern here is only with the topological fusion of multiple Bayes nets into a single Bayes net that allows information to flow as specified by *any* of the original networks. Section 2 does, however, review two crucial results from probability theory: First, probabilities compromised prior to the observation of any data are likely to have different implications from those compromised after data has been observed. Second, Bayes' rule implies that information in a Bayes net can flow in *either* direction (although not both at once). Thus, arc-reversal operations may be applied to a network without changing its qualitative relational structure. These results combine to imply that topological (or structural) fusion is both potentially useful and potentially achievable.

## 2   COMPROMISE AND CONSENSUS

Compromise and consensus are related terms that are often used to mean the same thing. Throughout this paper, however, we will attempt to use *compromise* to refer to the mathematical process of aggregating probability estimates and *consensus* to refer to the topological fusion of multiple Bayes nets into a single structure. The impact of both of these operations, of course, will be much the same. The consensus structure will contain compromised probabilities; conclusions based on the resultant Bayes net may or may not correspond to those that would have been reached by any of the contributing authors.

The mathematical aggregation of probabilities has long been discussed in the statistics and group decision-making literature. Although few of the results derived in this literature are relevant to structural fusion, they will become crucial in the design of compromised probabilities for the consensus network. For the sake of this paper, however, it should be safe to assume that probabilities will be compromised using the simplest aggregator, weighted average, which is not only conceptually straightforward, but also rather robust and surprisingly powerful (see [5] e.g.). This assumption clarifies the discussion without having much of an impact on any of our major results.

A question that is more relevant than *how* compromise probabilities can be derived, however, is *when* they should be derived. Raiffa outlined two reasonable positions, *prior compromise*, which is achieved before any data is observed, and *posterior compromise*, which is achieved afterwards. He further dis-cussed the relative merits of each position, and concluded that in many cases, prior compromises are preferable [9, pages 220-238]. To appreciate the difference between prior and posterior compromise, and to see the way that they can be reflected in a Bayes net, consider the example in Figure 1.

*Both authors agree on the relational structure:*

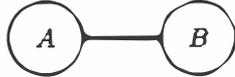

*Yet, they disagree on the assigned probabilities...*

| first author: | second author: |
|---|---|
| $P_1(A) = .80$ | $P_2(A) = .10$ |
| $P_1(B\|A) = .75$ | $P_2(B\|A) = .90$ |
| $P_1(B\|\bar{A}) = .10$ | $P_2(B\|\bar{A}) = .60$ |
| $P_1(A\|B) \approx .97$ | $P_2(A\|B) \approx .14$ |

$P^*(A|B) \approx \begin{cases} .56 & \text{prior compromise} \\ .66 & \text{posterior compromise} \end{cases}$

Figure 1: Prior and Posterior Compromise

In the example of Figure 1, the two authors *agree* about the relational structure, yet they *disagree* about the assigned probabilities. This example illustrates that prior and posterior compromises can lead to different results. If $B$ is observed before compromise is reached, the first author posits a posterior probability of $P_1(A|B) \approx .97$, while the second author determines that $P_2(A|B) \approx .14$. A posterior compromise that splits the difference (simple average) yields $P^*(A|B) \approx .56$. If compromise is reached (via averaging priors) before $B$ is observed, however, the compromise belief network calculates a posterior of $P^*(A|B) \approx .66$. The PMKBS architecture provided a structural framework within which posterior compromises among Bayes nets can be calculated [6, 7]. Topological fusion of Bayes nets will provide the structural framework for prior compromise.

There are two key concepts necessary to understand topological fusion: *graph union* and *arc reversal*. Graph union is rather straightforward. The union of two graphs consists of the unions of their node sets and their arc sets. The fundamental problem with using straight graph union for fusing Bayes nets is that it may generate cycles, thereby violating the topological constraints of the models. Fortunately, Bayes nets are not just DAGs; they are DAGs that model *information*, a quantity that can flow in either direction. This observation, captured mathematically by Bayes' rule, motivated researchers to introduce the arc-reversal operation, and to describe it as "information preserving" [10, 4]. The crux of an arc reversal is that if a Bayes net contains an arc from node $A$ to node $B$, the network's topology may be transformed by reversing the arc, (so that it now points from $B$ to $A$), and by adding arcs from each of $A$'s direct predecessors $(P(A))$ to $B$ and from each of $B$'s direct predecessors $(P(B))$ to $A$, *without affecting the network's underlying relational structure.*

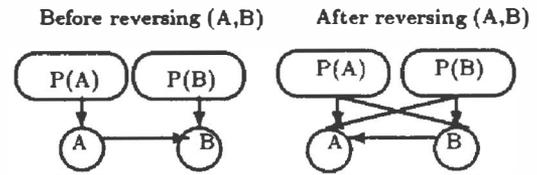

Figure 2: A reversal of arc $(A, B)$

The implication of arc reversal to topological fusion is obvious; if the structure derived through graph union contains cycles, some of the arcs need to be reversed. Indiscriminately applied arc reversals, however, may introduce new cycles even as they remove old ones. Thus, an algorithmic approach to graph-union-and-arc-reversal is required. Topological fusion, introduced in the next section, provides just such an approach.

## 3   TOPOLOGICAL FUSION

The *topological fusion* algorithm is based on *incremental graph union*. The algorithm's goal is to capture potential relations that are represented in a set of input Bayes nets by fusing them together into a single Bayes net. The DAG underlying the fused network begins with a complete set of nodes, (i.e., the union of all node sets), and the arcs from one input network. Arcs from a second network are then considered one at a time; they are classified as either (i) already in the fused network, (ii) includable in the fused network without causing any cycles, (iii) in need of reversal, or (iv) momentarily deferred. The treatment of arcs that belong to the first two categories is straightforward. The reversal of arcs in the third category, however, may force the introduction of new arcs into the second network. They, too, are considered, classified, and included in the fused network. The deferred arcs, (which were deferred because they pose certain potential topological difficulties) are then reconsidered, reclassified, and added to the fused DAG one at a time. When all of the second networks' arcs have been considered, a third network may be added, and so on. The algorithm is provably correct and tractable; it is guaranteed to terminate with a fused DAG that captures all necessary relations.

### 3.1   PRELIMINARIES

A formal statement and a proof of correctness of the topological fusion algorithm will require a bit of notation. Some of this notation is fairly standard in graph theory; some of it is new and specific to topological fusion.

- Let $D = (V, \vec{E})$ be a DAG that underlies a Bayes net.

- For each node $x \in V$, define the sets of $x$'s *direct predecessors and successors in DAG D*, $P_D(x)$ and $S_D(x)$ respectively:

$$P_D(x) = \{y \in V | (y, x) \in \vec{E}\}$$

$$S_D(x) = \{y \in V | (x, y) \in \vec{E}\}$$

Define (recursively) the set of *all of $x$'s successors* (direct and indirect) in *DAG D*, $S_D^*(x)$ as:

$$S_D^1(x) = S_D(x); \; S_D^i(x) = \bigcup_{y \in S_D^{i-1}(x)} S_D^1(y)$$

and

$$S_D^*(x) = \bigcup_{i=1}^{|V|} S_D^i(x).$$

Analogous sets may be defined for predecessors.

- A *topological value*, $\tau_D(x)$ over the nodes $x \in V$ in DAG $D$, is defined as the value assigned by a topological sort of $D$. Under this sort of partial ordering, all rooted nodes are assigned values of 0, and all other nodes are valued as the length of the *longest path* from a rooted node to them. Thus, topological values may be defined recursively as a function $\tau_D : V \rightarrow N$ by: $\tau_D(x) = 0$ if $P_D(x) = \emptyset$, and $\tau_D(x) = 1 + \max\{\tau_D(y) | y \in P_D(x)\}$ otherwise. Every DAG is therefore associated with an ordering, and as long as all arcs point from nodes with low topological values to nodes with high topological values, no cycles are possible. The addition of an arc that points from a high value node to a low value node *may* (but does not necessarily) introduce a cycle.

The definitions above were intentionally generic. A bit of specific machinery is necessary for the actual algorithm.

- Let $D_1 = (V_1, \vec{E}_1)$ and $D_2 = (V_2, \vec{E}_2)$ be two DAGs to be fused. (Assume without loss of generality that $|V_1| \geq |V_2|$).

- Let the fused DAG, denoted $D^* = (V^*, \vec{E^*})$, be initialized as $V^* = V_1 \cup V_2$ and $\vec{E^*} = \vec{E}_1$. This initialization begins $D^*$ with all of the nodes of $D_1$ and $D_2$, but only the arcs of $D_1$. As the algorithm proceeds, $D^*$ will grow by incrementally adding all of $D_2$'s arcs, or those that result from their reversal.

Next, define the three *sets* $DIR, REV$ and $EQ$ of arcs from $\vec{E}_2$, and the corresponding partial orderings $<_{REV}$ and $<_{EQ}$:

- $DIR$ contains arcs from $D_2$ that may be added directly to $D^*$, or those in which the topological relationships imposed by $D^*$ and $D_2$ agree:

$$DIR = \{(x, y) \in \vec{E}_2 | (x, y) \notin \vec{E^*}, \; \tau_{D^*}(x) < \tau_{D^*}(y)\}$$

$DIR$ need not be ordered.

- $REV$ contain arcs from $D_2$ that need to be reversed before they can be incorporated into $D^*$ without violating the topological constraints (this does not necessarily imply that if they were added, they would produce a cycle):

$$REV = \{(x, y) \in \vec{E}_2 | \tau_{D^*}(x) > \tau_{D^*}(y)\}$$

The relation $<_{REV}$ determines the priority for reversal. Haphazard arc reversals may produce cycles in $D_2$. A careful ordering of arcs to be reversed, however, precludes this possibility: $\forall (x_1, y_1), (x_2, y_2) \in REV, \; (x_1, y_1) <_{REV} (x_2, y_2)$ if $\tau_{D_2}(y_1) < \tau_{D_2}(y_2)$ or $(\tau_{D_2}(y_1) = \tau_{D_2}(y_2) \wedge \tau_{D_2}(x_1) \geq \tau_{D_2}(x_2))$. If $(x_1, y_1) <_{REV} (x_2, y_2)$, then $(x_1, y_1)$ is reversed before $(x_2, y_2)$. The importance of this ordering is discussed in the proof of Theorem 1.

- $EQ$ contains all arcs $(x, y) \in \vec{E}_2$ for which $\tau_{D^*}(x) = \tau_{D^*}(y)$ (and therefore, $(x, y) \notin \vec{E^*}$):

$$EQ = \{(x, y) \in \vec{E}_2 | \tau_{D^*}(x) = \tau_{D^*}(y)\}$$

The relation $<_{EQ}$ may then be defined on all $(x_1, y_1), (x_2, y_2) \in EQ$, as: $(x_1, y_1) <_{EQ} (x_2, y_2)$ if $\tau_{D^*}(x_1) > \tau_{D^*}(x_2)$ or $(\tau_{D^*}(x_1) = \tau_{D^*}(x_2) \wedge \tau_{D_2}(y_1) \geq \tau_{D_2}(y_2))$.

## 3.2  INFORMAL STATEMENT

An informal statement of the algorithm is now possible. The basic idea is to accept the topological ordering imposed by $D^*$, and to add all arcs from $D_2$ (or their reversals) in a manner that changes as few topological values as possible.

The first set considered is $REV$. Any arc $(x, y)$ selected from $REV$, reversed, and added (as $(y, x)$) to $D^*$, preserves the acyclicity of $D^*$ because if $(x, y) \in REV$, then $\tau_{D^*}(x) > \tau_{D^*}(y)$ and therefore $(y, x)$ does not change any topological values. Note that $(x, y)$ is also reversed in $D_2$. As a result, two new (possibly empty) sets of arcs $C_x = \{(z, x) | z \in P_{D_2}(y) \setminus P_{D_2}(x)\}$ and $C_y = \{(z, y) | z \in P_{D_2}(x) \setminus P_{D_2}(y)\}$ are generated in $D_2$. Arcs from these two sets must now be added to either $DIR$, $REV$, or $EQ$, and treated appropriately.

After $REV$ has been emptied (including any new arcs necessitated by arc reversals), attention can shift to the sets $EQ$ and $DIR$. Now, the arcs in $DIR$ do not appear in $D_1$, but they do point in the direction required by $\tau_{D^*}$, and may thus be added without changing the topological order. (It is not too difficult to see that they will not change any of the topological values imposed by $D^*$, either). The set $DIR$ may thus be emptied quickly; every arc in it may be added to $D^*$ directly.

The difficulty with $EQ$'s arcs lies in the impact that they have on $\tau_{D^*}$. Since an arc $(x, y)$ is in $EQ$ if and only if $\tau_{D^*}(x) = \tau_{D^*}(y)$, the addition of $(x, y)$ to $D^*$ *will* change the topological value of $y$ and $S^*_{D^*}(y)$; no nodes of lower topological value will be affected. Thus, if the first arc in $EQ$ selected for inclusion in $D^*$ was the one with the highest topological value ($\tau_{D^*}$), the only remaining arcs that may be affected are those among nodes of equal topological value. If, for example, $\tau_{D^*}(x) = \tau_{D^*}(y) = \tau_{D^*}(z)$ and by $<_{EQ}$, $(x, y)$ was selected as the next arc to be added to $D^*$, then by the definition of $<_{EQ}$ and the acyclicity assumption over $D_2$, *no* arcs of the form $(y, z)$ are in $EQ$ at all; hence, no arcs in $EQ$ need be recategorized as $REV$. Arcs of the form $(z, y)$ (if any), on the other hand, no longer belong in $EQ$; they are recategorized into $DIR$ and treated accordingly. The arc in $EQ$ with the next highest value may then be selected.

When all three sets have been emptied, $D_2$ has been converted to $D'_2$, a DAG that contains all of the same relevance relationships as $D_2$, but may look different because of arc reversals. $D^*$, which initially contained the nodes of $D_1$ and $D_2$ and the arcs of $D_1$, now also contains the arcs of $D'_2$, and thus represents the topological fusion of the two Bayes nets.

### 3.3   THE ALGORITHM

**algorithm FUSE_DAGS**

**INPUT**: Two DAGs $D_1 = (V_1, \vec{E_1})$, $D_2 = (V_2, \vec{E_2})$.
**OUTPUT**: DAGs $D^* = (V_1 \cup V_2, \vec{E^*})$ and $D'_2 = (V_2, \vec{E'_2})$ with the following properties:

1. There exists an "embedding" of $D_1$ and $D'_2$ in $D^*$.

2. $D'_2$ is obtained from $D_2$ by applying a valid sequence of *arc reversal* transformations over $D_2$.

Thus, $D^*$ contains "copies" of both $D_1$ and (albeit transformed), $D_2$.

**begin FUSE_DAGS**

1. initiate $D^* = (V_1 \cup V_2, \vec{E_1})$
2. initiate sets $DIR, REV$ and $EQ$ from $\vec{E_2}$
3. do until $REV = \emptyset$
4.    $(x, y) \leftarrow \min(REV)$ /* using $<_{REV}$. */
5.    $REV \leftarrow REV \setminus \{(x, y)\}$
6.    $\vec{E^*} \leftarrow \vec{E^*} \cup \{(y, x)\}$
   /* As a result of the pending arc reversal, */
   /* two new (possibly empty) sets of arcs */
   /* $C_x = \{(z, x) | z \in P_{D_2}(y) \setminus P_{D_2}(x)\}$ and */

/* $C_y = \{(z, y) | z \in P_{D_2}(x) \setminus P_{D_2}(y)\}$ are */
/* generated in $D_2$; their arcs should be */
/* distributed to the relevant sets. */
7.    $DIR \leftarrow DIR \cup \{(z, x) \in C_x | \tau_{D^*}(z) < \tau_{D^*}(x)\}$
         $\cup \{(z, y) \in C_y | \tau_{D^*}(z) < \tau_{D^*}(y)\}$
8.    $REV \leftarrow REV \cup \{(z, x) \in C_x | \tau_{D^*}(z) > \tau_{D^*}(x)\}$
         $\cup \{(z, y) \in C_y | \tau_{D^*}(z) > \tau_{D^*}(y)\}$
9.    $EQ \leftarrow EQ$  $\cup \{(z, x) \in C_x | \tau_{D^*}(z) = \tau_{D^*}(x)\}$
         $\cup \{(z, y) \in C_y | \tau_{D^*}(z) = \tau_{D^*}(y)\}$
10.   $\vec{E_2} \leftarrow \{\vec{E_2} \cup C_x \cup C_y \cup \{(y, x)\}\} \setminus \{(x, y)\}$
11. **enddo**
12. do until $EQ = DIR = \emptyset$
13.   $\vec{E^*} \leftarrow \vec{E^*} \cup DIR$
14.   $DIR \leftarrow \emptyset$
    /* $DIR$ is empty when $EQ$ is considered. */
15.   $(x, y) \leftarrow \min(EQ)$ /* using $<_{EQ}$. */
16.   $EQ \leftarrow EQ \setminus \{(x, y)\}$
17.   $\vec{E^*} \leftarrow \vec{E^*} \cup \{(x, y)\}$
    /* When this arc was added to $\vec{E^*}$, it */
    /* changed $\tau_{D^*}(y)$. Hence, for all arcs */
    /* $(z, y) \in EQ$, $\tau_{D^*}(z) = \tau_{D^*}(y) - 1$; */
    /* these arcs are transferred to $DIR$. */
18.   $DIR \leftarrow \{(z, y) | (z, y) \in EQ\}$
19.   $EQ \leftarrow EQ \setminus DIR$
20. **enddo**
**end FUSE_DAGS**

### 3.4   PROOF OF CORRECTNESS

The thrust of this section is a proof of correctness of algorithm **FUSE_DAGS**. Our intention is not to prove that the consensus structure that it generates is unique (it is not) or optimal (a yet undefined term), but rather that the algorithm works as it should. In other words: (i) the algorithm halts, (ii) $D'_2$ is obtained from $D_2$ by a *valid* sequence of arc reversals (i.e., no cycles are formed), and (iii) the resultant $D^*$ generated by the algorithm contains embeddings of both $D_1$ and $D'_2$. This is an important property: since the consensus structure contains "copies" of $D_1$ and (albeit transformed) $D_2$, projections onto appropriate sets of nodes and arcs reconstruct each individual author's model.

**Lemma 1** *The following properties hold at all times for all arcs $(x, y) \in D_2$:*
*If $(x, y) \in DIR$ then $\tau_{D^*}(x) < \tau_{D^*}(y)$.*
*If $(x, y) \in REV$ then $\tau_{D^*}(x) > \tau_{D^*}(y)$.*
*If $(x, y) \in EQ$, $\tau_{D^*}(x) = \tau_{D^*}(y)$.*
*When $(x, y)$, however, is removed from $EQ$ and added to $D^*$, topological values are changed such that $\tau_{D^*}(y) = \tau_{D^*}(x) + 1$ and hence, some of these relationships may be temporarily violated during the execution of lines (17-18).*

**Lemma 2** *At any time, if an arc $(x, y) \in \vec{E_2}$ but $(x, y) \notin \vec{E^*}$, then $(x, y)$ is in either $DIR, REV$ or*

*EQ.*

These lemmas are trivial (and almost definitional). Formal proofs are left as an exercise for particularly motivated readers.

**Theorem 1** *The construction specified by algorithm* **FUSE_DAGS** *creates no cycles in either $D^*$ or $D_2$ during any iteration of the two do loops. It thus specifies a valid sequence of arc reversals in the transformation of $D_2$ into $D_2'$.*

*Proof:*

Since $D_1$ is acyclic by definition, so is the initialized $D^*$. Now, assume that a cycle(s) is formed in $D^*$ during some iteration of any of the do loops. Consider the *first* time that a cycle was formed, (i.e., $D^*$ was acyclic until the addition of arc $(x, y)$ introduced a cycle). According to lemma (1), $\tau_{D^*}(x) \leq \tau_{D^*}(y)$ when $(x, y)$ was added to $\vec{E^*}$. If $(x, y)$ formed a cycle, however, there must have already been a directed path in $D^*$, $y = z_1, \ldots, z_n = x$, $n \geq 1$. By the definition of $\tau_{D^*}$, then, $\tau_{D^*}(z_1) < \cdots < \tau_{D^*}(z_n)$, or (by transitivity), $\tau_{D^*}(y) < \tau_{D^*}(x)$. This contradiction proves that no cycle can ever be introduced into $D^*$.

Next, assume that a cycle was formed in $D_2$, and consider once again the *first* time that the cycle was introduced. Since the original $D_2$ was acyclic, the cycle must have been formed during the reversal of arc $(x, y) \in REV$ (line (10)). Moreover, $(y, x)$ must thus be in at least one of the cycles formed as a result of reversing $(x, y)$ (it is, for example, in the longest such a cycle). Let this resultant cycle from $y$ to $y$ be denoted $(y, x = z_1), \ldots, (z_{n-1}, z_n = y)$ where $n \geq 1$ and $(y, x)$ is $(x, y)$ reversed. (Assume the cycle is of length 3 or more, the case of length 2 is trivial). The topological ordering imposed by $D_2$ on the nodes in this cycle *before* $(x, y)$ is reversed, then, is $\tau_{D_2}(x = z_1) < \ldots < \tau_{D_2}(z_{n-1}) < \tau_{D_2}(z_n = y)$. Now consider the set $REV$ when $(x, y)$ was selected as $\min(REV)$ (line (4)). If $(x = z_1, z_2)$ had also been in $REV$ at the time (as a potential candidate for reversal), then *it* would have been selected and deleted from $REV$ *before* $(x, y)$ because $\tau_{D_2}(z_2) < \tau_{D_2}(z_3) < \cdots < \tau_{D_2}(z_n = y)$ (and by transitivity, $\tau_{D_2}(z_2) < \tau_{D_2}(z_n = y)$). The definition of $<_{REV}$ indicates that $(x, z_2) <_{REV} (x, y)$ and $(x, z_2)$ would thus have been selected before $(x, y)$. The arc $(x = z_1, z_2)$ is thus *not* in $REV$; it must therefore be either in $D^*$, in $DIR$, or in $EQ$ (by lemma (2)). Similar arguments may be applied, in turn, to each of the arcs in the cycle, to show that none of them can be in $REV$. All of these arcs must thus be in either $D^*$, $DIR$, or $EQ$ and regardless of each of these arcs' classification, then, lemma (1) indicates that $\forall i, 1 \leq i < n, \tau_{D^*}(z_i) \leq \tau_{D^*}(z_{i+1})$ (and by transitivity $\tau_{D^*}(x = z_1) \leq \tau_{D^*}(z_n = y)$). According to lemma (1), however, if $(x, y)$ is in $REV$ then

$\tau_{D^*}(x) > \tau_{D^*}(y)$, thereby producing a contradiction. This contradiction proves that $D_2$ is acyclic at the end of every iteration of any of the do loops. Thus, the sequence of reversals is *valid* because no cycles were formed at any point.    □

Theorem 1 proves that the fusion of $D_1$ and $D_2$ into $D^*$ is valid *when the algorithm terminates*. It does not, however, prove that the algorithm must terminate (or in fact, that it will ever terminate). It also says nothing about the algorithm's complexity. These items are addressed by Theorem 2.

**Lemma 3** *At any given time, each arc $(x, y) \in V_2 \bigotimes V_2$ (i.e., any potential arc of $D_2'$) can be in at most one of $DIR, REV$ or $EQ$ (with the technical exception of the time lines (17) and (18) are executed). In other words, $DIR, REV$ and $EQ$ are always pairwise disjoint, and if an arc $(x, y)$ is in one of the sets, then neither it nor its reversal, $(y, x)$, is in either of the other two.*

**Lemma 4** *At any time, if an arc $(x, y)$ is added to $\vec{E^*}$ (and thus deleted from its corresponding set), then at any subsequent iteration of each of the two do loops in the algorithm, $(x, y), (y, x) \notin DIR \cup REV \cup EQ$*

**Theorem 2** *Algorithm* **FUSE_DAGS** *terminates if both $V_1$ and $V_2$ are finite; its complexity is $O(\max\{|V_1|, |E_1|, |V_2|^3\})$.*

*Proof:*

The number of iterations of the algorithm's two loops must be finite, because:

1. At most one occurrence of a given $(x, y) \in V_2 \bigotimes V_2$ is found in sets $DIR, REV$ and $EQ$ at any given time (by lemma (3)).

2. Once an arc $(x, y)$ is deleted from $DIR, REV$, or $EQ$, and added to $\vec{E^*}$, neither $(x, y)$ nor $(y, x)$ can ever again appear in any of the sets (by lemma (4)).

3. During each iteration of each of the two do loops, *at least* one arc is deleted from one of the sets and added to $\vec{E^*}$.

Now, let $K_{REV}$ denote the number of iterations of the loop "do until $REV = \emptyset$" (lines (3-11)), and let $K_{EQ}$ denote the number of iterations of the loop "do until $EQ = DIR = \emptyset$" (lines (12-20)). Then it is readily apparent that $K_{REV} + K_{EQ} \leq |V_2| \cdot (|V_2| - 1)/2$ and that $K_{EQ} \leq |V_2|$. Since each of these iterations is clearly finite, the algorithm terminates.

Next, consider the algorithm's preprocessing stages. Initiating $DIR, REV$ and $EQ$ requires an evaluation of $\tau_{D^*}(x)$ for each $x \in V_1$ (nodes $y$ added from $V_2$ where $y \notin V_1$ are immediately assigned $\tau_{D^*}(y) = 0$). Such an evaluation can be done by a topological sort on $D_1$, where $D_1$ is represented by an adjacency matrix and $\tau_{D^*}$ is evaluated recursively by its definition. This preprocessing thus takes $O(\max\{|E_1|, |V_1|\})$ steps using a simple *depth-first search (dfs)*. A second *dfs* performed on $D_2$, initiates the three sets in $O(\max\{|E_2|, |V_2|\})$ steps.

The complexity of the two **do** loops and individual operations must also be evaluated. The evaluation of $\min(REV)$ requires $O(\log|REV|)$ steps (assuming $REV$ is maintained as a heap; constructing such a heap takes $O(|REV|)$ steps). Arc reversal, however, also necessitates a bit of bookkeeping. The identification of the newly generated arcs, and the updating of the topological values imposed by $D_2$ on $x, y, S_{D_2}^*(x)$, and $S_{D_2}^*(y)$, are a main source of the algorithm's overall complexity because reversing an arc $(x, y) \in REV$ forces an examination of $P_{D_2}(x)$ and $P_{D_2}(y)$. Both of these immediate-predecessor sets may be of size $O(|V_2|)$, and updating topological values requires an examination of $S_{D_2}^*(x)$ and $S_{D_2}^*(y)$, (both may also be of size $O(|V_2|)$). Since a complete graph contains $|V_2| \cdot (|V_2| - 1)/2$ arcs, the iterations of this loop could take as many as $O(|V_2|^3)$ steps in the worst case (i.e., if all arcs in a complete graph had to be reversed).

The analysis of arcs initially placed in $EQ$ is a bit more complex; it requires a preliminary claim (which is almost trivially proven):

**Claim 1** *If an arc $(x, y)$ is added to $\vec{E^*}$ from $EQ$ during some iteration $i > 0$ of the loop "do until $EQ = DIR = \emptyset$", then during any subsequent iteration $j > i$ (or, for that matter, at any subsequent step of iteration $i$), no arc $(u, v)$, $u, v \in V_2$ for which $\tau_{D^*}(u) \le \tau_{D^*}(y)$ or $\tau_{D^*}(v) \le \tau_{D^*}(y)$ is considered as part of sets $EQ$ or $DIR$ (and obviously, $REV$).*

This claim indicates that once $(x, y)$ is added to $\vec{E^*}$ (thereby changing the topological values of $y$ and its successors in $D^*$), the topological values of $y$'s successors need not be updated but *only $y$'s* value requires an update, which takes a constant number of steps. The upper bound on the number of steps required to add the arcs from $EQ$ (and at most $K_{EQ}$ such arcs are added to $\vec{E^*}$ directly from $EQ$), is thus $K_{EQ} d_{EQ} |V_2| \log |V_2|$ for some $d_{EQ} > 1$, or $O(|V_2|^2 \log |V_2|)$ steps; note that finding $\min(EQ)$ (by $<_{EQ}$) requires only $O(\log |V_2|)$ steps.

These analyses of the algorithm's components combine to prove that **FUSE_DAGS** takes $O(\max\{|V_1|, |E_1|, |V_2|^3\})$ steps.    □

Finally, define the following sets of arcs:

$$S_1 = \{(x, y) | (x, y) \in \vec{E_1} \setminus \vec{E_2'}\}$$

$$S_2 = \{(x, y) | (x, y) \in \vec{E_2'} \setminus \vec{E_1}\}$$

$$S_3 = \{(x, y) | (x, y) \in \vec{E_1} \cap \vec{E_2'}\}$$

$E_2'$ is the set of arcs of $D_2'$, where $D_2'$ was obtained from $D_2$ by a valid sequence of arc reversal transformations (theorem (1)). Note too that $\vec{E^*} = \vec{E_1} \cup \vec{E_2'}$; otherwise the algorithm would not have terminated. To see the embedding of $D_1$ and $D_2'$ in $D^*$, project $D^*$ on the relevant sets of nodes and arcs, to yield $D^*|_{V_1, S_1 \cup S_3} = D_1$ and $D^*|_{V_2, S_2 \cup S_3} = D_2'$, as required.

## 4    AN EXAMPLE

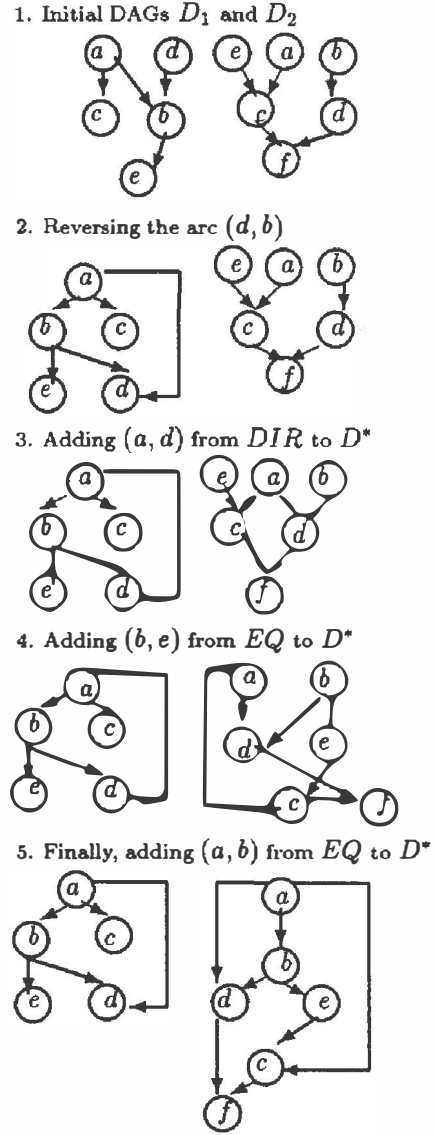

Figure 3: Fusing DAGs $D_1$ and $D_2$.

The previous few sections introduced an algorithm for topological fusion, and proved that it is correct and efficient. Figure 3 gives a simple ex-ample of how this algorithm can be used to fuse two DAGs, $D_1 = (V_1 = \{a, b, c, d, e, f\}, \vec{E}_1 = \{(a, c), (b, d), (e, c), (c, f), (d, f)\})$ and $D_2 = (V_2 = \{a, b, c, d, e\}, \vec{E}_2 = \{(a, b), (a, c), (d, b), (b, e)\})$.

The fusion example of Figure 3 ($D_2$ is on the left, $D^*$ on the right) proceeds as follows. In step 1, two DAGs ($D_1$ and $D_2$) are given as input to the algorithm. The fused graph and arc sets may be initialized as $D^* = D_1, DIR = \emptyset, REV = \{(d, b)\}$ and $EQ = \{(a, b), (b, e)\}$. The loop "do until $REV = \emptyset$" is then executed. Arc $(d, b) = \min(REV)$ is selected for reversal in step 2, thereby generating arc $(a, d)$. Since $\tau_{D^*}(a) < \tau_{D^*}(d), (a, d) \in DIR$. This addition to $DIR$ shifts control to the loop "do until $EQ = DIR = \emptyset$." Control then enters the loop, and $(a, d)$ is added directly to $D^*$ (line (13)), as shown in step 3. Then, $DIR = \emptyset$ and $EQ = \{(a, b), (b, e)\}$, line 15 is executed, and since $\tau_{D^*}(a) = \tau_{D^*}(b) = \tau_{D^*}(e) = 0$ but $\tau_{D_2}(b) = 1 < \tau_{D_2}(e) = 2$, $(b, e)$ is selected as $\min(EQ)$. Step 4 results from the selection and addition of $(b, e)$. As a result of adding $(b, e)$ to $D^*$, $\tau_{D^*}(e) = 1$. This change forces *no* transfer of arcs from $EQ$ to $DIR$. A new iteration of the loop "do until $EQ = DIR = \emptyset$" then begins. $DIR = \emptyset$, but $EQ = \{(a, b)\}$, and thus arc $(a, b)$ is added to $D^*$ in step 5 (line 15). Since $DIR = REV = EQ = \emptyset$, **FUSE_DAGS** terminates.

## 5   SUMMARY

Bayes nets, belief networks, and influence diagrams are relatively recent innovations. As a result, most of their development to date has been on the simplest class of *single-author* models. The introduction of more sophisticated settings, such as multiple modelers and/or distributed system-design efforts, raise a variety of interesting questions. One such question involves the nature of compromise and consensus. If two models (or contributors) agree about some things, but disagree about others, what sort of compromise is possible? Two answers come to mind: prior compromise and posterior compromise. Posterior compromises are the more obvious of the two. They let each model (contributor) process all data to arrive at an independent response, and then split the difference. Prior compromises, on the other hand, force compromise to be reached on all points *before* data is observed.

This paper began the discussion of prior compromises in a Bayes net setting. It outlined the problem, and developed an algorithm, **FUSE_DAGS**, which produces the topological fusion necessary for prior compromise. The algorithm's meaningfulness, of course, is restricted to Bayes nets; arc reversals are not defined on other DAGs; and they certainly do not preserve all types of relationships. Probabilities associated with the nodes of $D^*$ may be aggregated by any standard aggregation function, including (but not restricted to) weighted average. The method used in the algorithm's construction guarantees that all necessary input probabilities were available when needed. Further discussions of the relationship between topological fusion and aggregated compromise, however, must be relegated to the future. This paper merely laid the graph-theoretic groundwork necessary to discuss the use of Bayes nets as models of prior compromise, and perhaps more importantly, it introduced the issue to the community of Bayes net researchers. Work on this topic has only just begun. A great deal more must follow.